\documentclass[letterpaper, 10 pt, journal, twoside]{IEEEtran} 

\IEEEoverridecommandlockouts                              

\usepackage{amsmath}
\usepackage{amssymb}
\usepackage{mathtools}
\usepackage{graphicx}
\usepackage{float}
\usepackage{array}
\usepackage{tikz}
\usepackage{listings}
\usepackage{hyperref}
\usepackage{graphicx}
\usepackage{cite}
\usepackage{dsfont}
\usepackage{mathtools,etoolbox}
\usepackage[ruled, linesnumbered]{algorithm2e}
\usepackage[none]{hyphenat}
\usepackage[utf8]{inputenc}
\usepackage[english]{babel}

\usepackage{amsthm}
\usepackage{xfrac}
\usepackage{nicefrac}
\usepackage{booktabs}
\usepackage[switch, pagewise]{lineno}
\usepackage{nicefrac}
\usepackage{flushend}

\newtheorem*{remark}{Remark}
\DeclareMathOperator*{\argmin}{argmin} 
\DeclareMathOperator*{\argmax}{argmax} 
\newcommand\abs[1]{\left|#1\right|}
\hypersetup{colorlinks=true,urlcolor=blue,citecolor=red}

\newcommand{\norm}[1]{\left\lVert#1\right\rVert} 
\title{GapFlyt: Active Vision Based Minimalist Structure-less Gap Detection For Quadrotor Flight}
\author{Nitin J. Sanket, Chahat Deep Singh, Kanishka Ganguly, Cornelia Ferm{\"u}ller, Yiannis Aloimonos 
\thanks{Manuscript received February 15, 2018; accepted May 19, 2018. Date of
publication June 4, 2018; date of current version June 15, 2018. This letter was
recommended for publication by Associate Editor V. Lippiello and Editor J.
Roberts upon evaluation of the reviewers’ comments. This work was supported
in part by the Brin Family Foundation, in part by the Northrop Grumman
Corporation, and in part by the National Science Foundation under Grants
SMA 1540917 and CNS 1544797, respectively. \textit{(Nitin J. Sanket and Chahat
Deep Singh contributed equally to this work.) (Corresponding author: Nitin J.
Sanket.)}}
\thanks{All authors are associated with University of Maryland Institute for Advanced Computer Studies, College Park. Emails: \{{\tt\footnotesize nitin, chahat, kganguly, fer, yiannis}\} {\tt \footnotesize @umiacs.umd.edu}}
\thanks{Digital Object Identifier (DOI): see top of this page.}
}

\begin{document}
\makeatletter
\g@addto@macro\@maketitle{
\begin{figure}[H]
   \setlength{\linewidth}{\textwidth}
   \setlength{\hsize}{\textwidth}
    \centering
    \includegraphics[width=\textwidth]{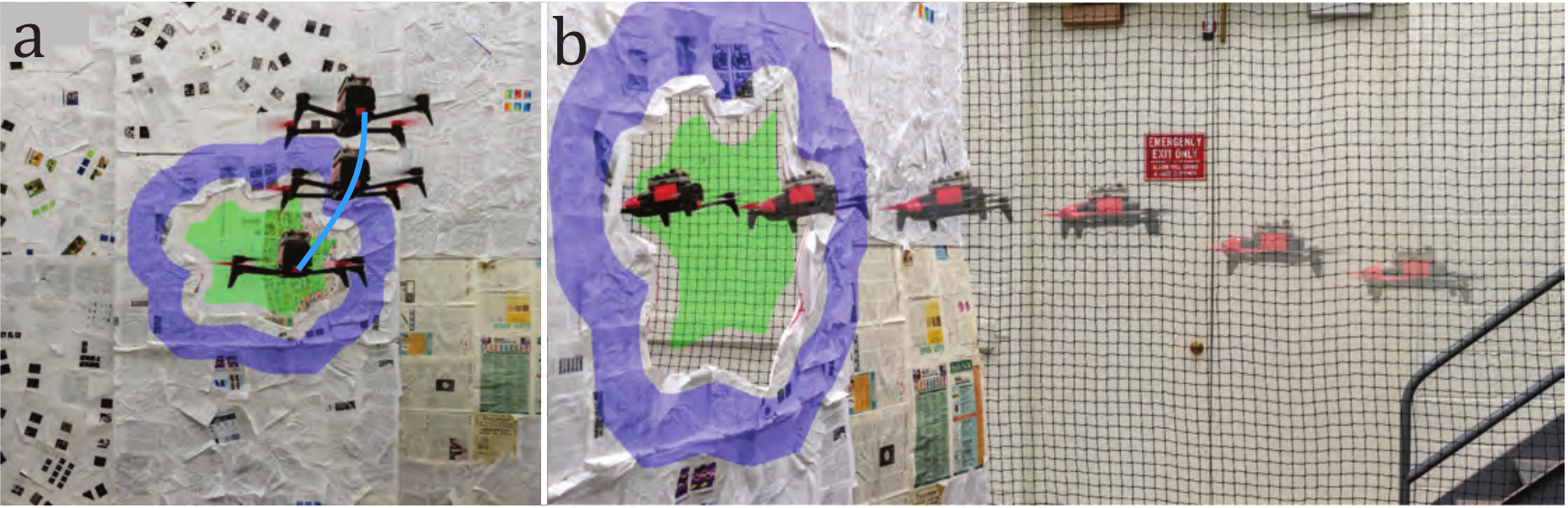}
    \caption{Different parts of the pipeline: (a) Detection of the unknown gap using active vision and TS$^2$P algorithm (cyan highlight shows the path followed for obtaining multiple images for detection), (b) Sequence of quadrotor passing through the unknown gap using visual servoing based control. The blue and green highlights represent the tracked foreground and background regions respectively. Best viewed in color.}
    \label{fig:Overview}
    \end{figure}
}
\maketitle

\setcounter{figure}{1}

\begin{abstract}

Although quadrotors, and aerial robots in general, are inherently active agents, their perceptual capabilities in literature so far have been mostly passive in nature. Researchers and practitioners today use traditional computer vision algorithms with the aim of building a representation of general applicability: a 3D reconstruction of the scene. Using this representation,  planning tasks are constructed and accomplished to allow the quadrotor to demonstrate autonomous behavior. These methods are inefficient as they are not task driven and such methodologies are not  utilized by flying insects and birds. Such agents have been solving the problem of navigation and complex control for ages without the need to build a 3D map and are highly task driven.\\
\indent In this paper, we propose this framework of bio-inspired perceptual design for quadrotors. We use this philosophy to  design a minimalist sensori-motor framework for a quadrotor to fly though unknown gaps without an explicit 3D reconstruction of the scene using only a monocular camera and onboard sensing. We successfully evaluate and demonstrate the proposed approach in many real-world experiments with different settings and window shapes, achieving a success rate of 85\% at 2.5ms$^{-1}$ even with a minimum tolerance of just 5cm. To our knowledge, this is the first paper which addresses the problem of gap detection of an unknown shape and location with a monocular camera and onboard sensing.
\end{abstract}

\textbf{\textit{\small{Keywords -- Active Vision, Gap Detection, Quadrotor, Visual
Servoing, Deep Learning in Robotics and Automation, Optical
Flow, Tracking, Collision Avoidance, Computer Vision for
Automation, Aerial Systems: Perception and Autonomy, Visual
Tracking and Visual-Based Navigation.}}}

\section*{Supplementary Material}
The supplementary hardware tutorial, appendix, code and
video are available at \url{prg.cs.umd.edu/GapFlyt.html}.

\section{Introduction and Philosophy}
\IEEEPARstart{T}{he} quest to develop intelligent and autonomous quadrotors \cite{Exploration, Inspection, SearchAndRescue} has taken a center stage in the recent years due to their usefulness in aiding safety and intuitive control. To achieve any form of autonomy, quadrotors need to have perceptual capabilities in order to sense the world and react accordingly. A generic and fairly common solution to providing perceptual capabilities is to acquire a 3D model of its environment. Creating such a model is very useful because of its general applicability -- one could accomplish many tasks using the same model. The process of obtaining a 3D model of the environment using onboard sensing and a myriad of different sensors has gained momentum in the last few years \cite{SLAMSurvey}. Sensors like the LIDAR, RGB-D and stereo camera cannot be used on a small quadrotor due to their size, weight, and power limitations. This constrains us to a monocular camera along with the already present onboard inertial sensor (IMU) and many algorithms have been developed based on the same principle \cite{VINS}\cite{ROVIO}.

\begin{table*}[t!]
\centering
\caption{Minimalist design of autonomous quadrotor (drone) behaviours.}
\resizebox{\textwidth}{!}{
\label{tab:MinimalistPhiloshopy}
\begin{tabular}{lll}
\toprule
Competence & Passive Approach & Active and Task-based Approach \\
 \hline \\
 Kinetic stabilization & Optimization of optical flow fields & Sensor fusion between optical flow and IMU measurements \\
 Obstacle avoidance & Obtain 3D model and plan accordingly & Obtain flow fields and extract relevant information from them \\
 Segmentation of independently moving objects & Optimization of flow fields & Fixation and tracking allows detection \\
 Homing & Application of SLAM & Learn paths to home from many locations \\
 Landing & Reconstruct 3D model and plan accordingly & Perform servoing of landing area and plan appropriate policy \\
 Pursuit and Avoidance & Reconstruct 3D model and plan accordingly & Track while in motion \\
 Integration: Switching between behaviors & Easy: The planner interacts with the 3D model & Hard: An attention mechanism on ideas switching between behaviors\\
 \bottomrule\\[-20pt]
\end{tabular}}
\end{table*}

Instead of attempting to recover a 3D model of the scene, we want to recover a ``minimal'' amount of information that is sufficient to complete the task under consideration. We conceptualize an autonomous quadrotor as a collection of processes that allow the agent to perform a number of behaviors (or tasks) (Table \ref{tab:MinimalistPhiloshopy}). At the bottom of the hierarchy is the task of kinetic stabilization (or egomotion). Next comes the ability for obstacle avoidance, where the obstacles could be static or dynamic, followed by ability to perform homing, i.e., to find a specific location in an environment. Last in the hierarchy comes the ability to land (on a static or a dynamic surface) and the ability to pursue, or escape from, another agent. This hierarchy of competences, with each competence needing the one before it, constitutes the sensorimotor system of the quadrotor agent. These behaviors can be accomplished without an explicit 3D reconstruction because of the ``active'' nature of the quadrotor. The quadrotor can perform maneuvers and control the image acquisition process, thus introducing new constraints that were not there before - this is called ``active'' vision \cite{ActiveVision, SukhtameActive, BajcsyActive}. This methodology was inspired by the fruit fly \cite{FruitFly}. Prior research has shown that fruit flies, and other insects \cite{ExploratoryMovement} \cite{LocustsAndMantids}, do not perceive depth directly. It is achieved by a balance of active and exploratory procedures. In this paper, we focus on the second competence of obstacle avoidance. Specifically, the question this paper deals with is: \textit{``Can a quadrotor manage to go through an arbitrarily shaped gap without building an explicit 3D model of a scene, using only a monocular camera?''} We develop the visual mathematics of the solution, evaluate and demonstrate the proposed approach in many real experiments with different settings and window shapes.

Traditional computer vision based approaches such as sparse or dense reconstruction \cite{GeoSLAM, RGBDSLAM, SLAMDunk, DSO} have been used to obtain a 3D structure of the scene over which sophisticated path planners  have been used to plan collision free paths. Lately, deep-learning driven approaches have taken a center stage in solving the problem of fast collision avoidance and safe planning on quadrotors \cite{TrailPaper}. Most of these neural network approaches compute Fast Fourier Transforms (FFTs) for large filter sizes \cite{FFT}. Such approaches can be processed on a Field-Programmable Gate Array (FPGA), rather than a Graphical Processing Unit (GPU) to drastically improve the computation performance and power efficiency \cite{FPGA1}\cite{FPGA2}.

To our knowledge, this is the first paper which addresses the problem of gap detection with a monocular camera and onboard sensing. However, the problem of going through gaps has fascinated researchers from many years and some of the recent works which present algorithms for planning and control can be found in \cite{KumarWindow} \cite{ScaramuzzaWindow}. Some of the works which paved way to the bio-inspired approach used in this paper can be found in \cite{franceschini1992insect, srinivasan1999robot, serres2017optic, scheper2016behavior}.

The key contributions of this paper are given below:
\begin{itemize}
\item Active vision based structure-less minimalist gap detection algorithm -- Temporally Stacked Spatial Parallax or TS$^2$P (Fig. \ref{fig:Overview}a).
\item Visual servoing on a contour for the quadrotor to fly through unknown gaps (Fig. \ref{fig:Overview}b).
\end{itemize}
\subsection{Organization of the paper:}
Sec. \ref{sec:TS2P} presents the detection algorithm of a gap in an arbitrary shaped window using Deep Learning based optical flow \textit{and the role of active vision in such algorithms.} Sec. \ref{sec:VisualServo} describes the algorithm used for tracking the gap/safe point and the quadrotor control policy. Sec. \ref{sec:Expts} illustrates the experimental setup and provides error analyses and comparisons with traditional approaches. We finally conclude the paper in Sec. \ref{sec:Conc} with parting thoughts on future work.

\subsection{Problem Formulation and Proposed Solutions}
\begin{figure}[b!]
    \centering
    \includegraphics[width=\columnwidth]{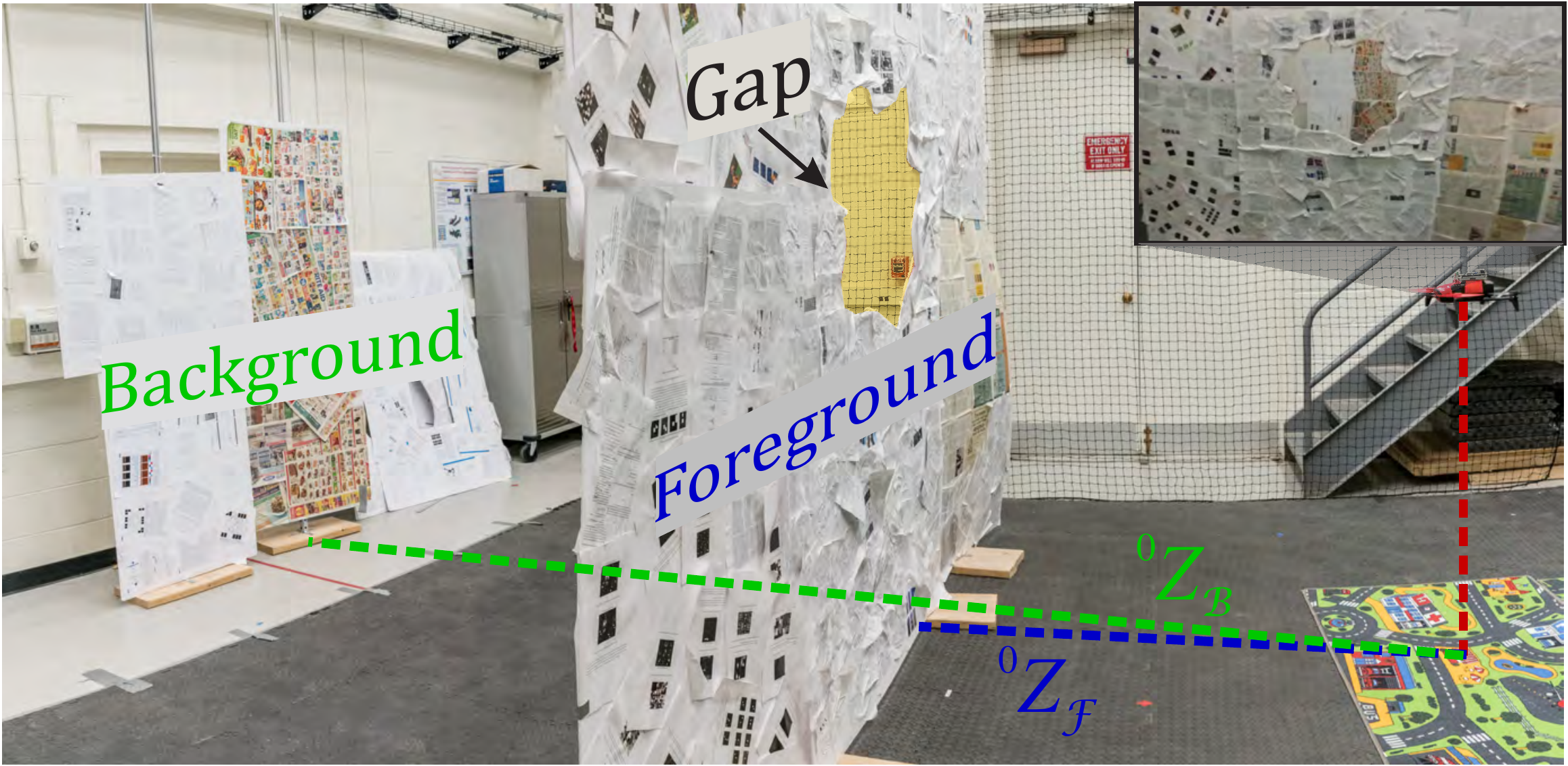}
    \caption{Components of the environment. On-set Image: Quadrotor view of the scene.}
    \label{fig:FgBgDist}
\end{figure}

A quadrotor is present in a static scene (Fig. \ref{fig:FgBgDist}), where the absolute depth of each point as `seen' from the camera can be modelled as an univariate bimodal gaussian distribution. The location at which there is a maximum spatial depth discrepancy between pixels (projected point) is defined as the Contour ($\mathcal{C}$) of the opening. In this paper, the words boundary, gap, window or opening refer to the same entity and will be used interchangeably. Any pixel close to the mode corresponding to the lower depth value is defined as the Foreground ($\mathcal{F}$) and similarly that corresponding to the higher depth value is defined as the Background ($\mathcal{B}$). A simple way of solving the problem of finding the gap for a near fronto-parallel pose is to find the depth discrepancy which is a trivial clustering problem when the depth is known. The depth can be known if multiple calibrated cameras are used or the entire scene is reconstructed in 3D. These methods are computationally expensive \cite{SLAMSurvey}. In this paper, we propose a `minimalist' solution to find any arbitrary shaped gap for the quadrotor to go through using Temporally Stacked Spatial Parallax (TS$^2$P) algorithm. A control policy based on contour visual servoing is used to fly through unknown gaps.

\section{Gap Detection using TS$^2$P}
\label{sec:TS2P}
\begin{figure}[t!]
    \centering
    \includegraphics[width=0.6\columnwidth]{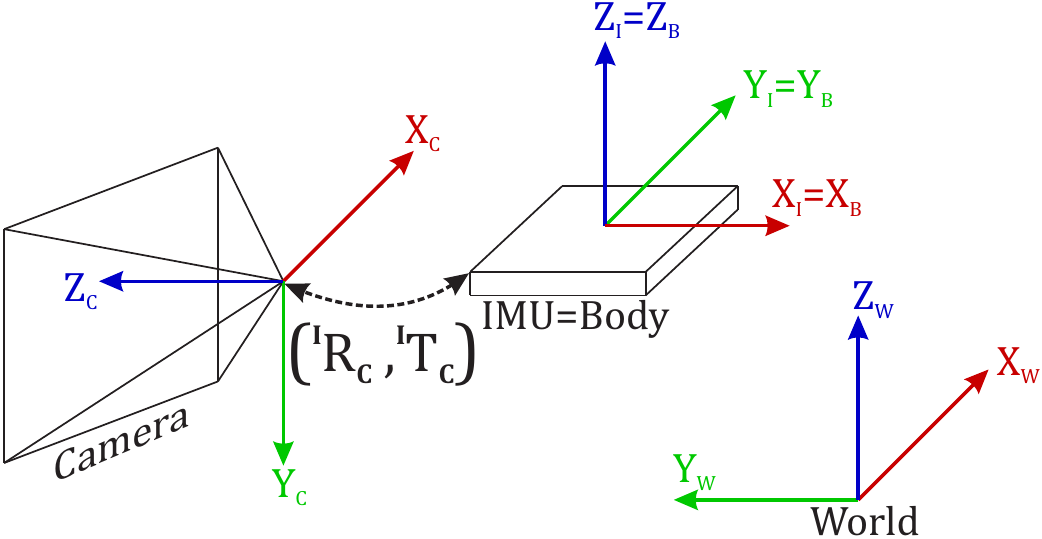}
    \caption{Representation of co-ordinate frames.}
    \label{fig:CoordinateFrames}
\end{figure}

    Before we explain the procedure for gap detection, we need to formally define the notation used. $\left(\prescript{a}{}{X}_b,\prescript{a}{}{Y}_b,\prescript{a}{}{Z}_b\right)$ denotes the coordinate frame of $b$ represented in the reference of $a$. The letters $I$, $C$, $B$ and $W$ are used as sub/superscript to denote quantities related to Inertial Measurement Unit (IMU), Camera, Body and World respectively. If a sub/superscript is omitted, the quantity can be assumed to be in $W$. $C$ and $I$ are assumed to be rigidly attached to each other with known intrinsics and extrinsics. $B$ is defined to be aligned with $I$ (Fig. \ref{fig:CoordinateFrames}). A pinhole camera model is used for the formation of the image. The world point $\mathbf{X}$ gets projected onto the image plane point $\mathbf{x}$.

The camera captures images/frames at a certain frequency. Let the frame at time $t^i$ be denoted as $\mathbb{F}^i$ and is called the $i^{\text{th}}$ frame. Optical flow \cite{MotionField} between $i^{\text{th}}$ and $j^{\text{th}}$ frames is denoted by $\prescript{j}{i}{\dot{p}_{\mathbf{x}}}$ which is a sum of both translational $\left(\prescript{j}{i}{\dot{p}_{\mathbf{x}, T}}\right)$ and rotational $\left(\prescript{j}{i}{\dot{p}_{\mathbf{x}, R}}\right)$ components and are given by: 

\[\resizebox{\columnwidth}{!}{
$\prescript{j}{i}{\dot{p}_{\mathbf{x}, T}} = \dfrac{1}{Z_\mathbf{x}}
    \begin{bmatrix}
    x V_z - V_x \\
    y V_z - V_y \\
    \end{bmatrix}; \,\, \prescript{j}{i}{\dot{p}_{\mathbf{x}, R}} = \begin{bmatrix}
    xy & -(1+x^2) & y \\
    (1+y^2) & -xy & -x \\
    \end{bmatrix} \Omega $}
\]

where $\mathbf{x}=\begin{bmatrix} x & y \end{bmatrix}^T$ denotes the pixel coordinates in the image plane, $Z_\mathbf{x}$ is the depth of  $\mathbf{X}$ (world point corresponding to pixel $\mathbf{x}$), $V = \begin{bmatrix} V_x & V_y & V_z \end{bmatrix}^T$ and  $\Omega$ are the linear and angular velocities of the $C$ in $W$ between $t^i$ and $t^j$ respectively.

The otherwise complex depth estimation problem can be trivialized by moving in a certain way\cite{ActiveVision}. These ``active'' vision principles dictate us to control the quadrotor's movements so as to make the interpretation of optical flow easier. Since the quadrotor is a differentially flat system \cite{DifferentialFlatness}, the rotation about $\left(X_B, Y_B, Z_B\right)$ or roll, pitch and yaw can be decoupled. As an implication, the movements can be controlled in a way to simplify the depth $\left( Z_\textbf{x}\right)$ estimation problem. The quadrotor is controlled in such a way that $\Omega\approx0$, $V_z \ll V_x$ and  $V_z \ll V_y$, then the optical flow can be modelled as:

\[\prescript{j}{i}{\dot{p}_{\mathbf{x}}}=
\prescript{j}{i}{Z_{\mathbf{x}}}^{-1}\begin{bmatrix}
    -\prescript{j}{i}{V_x} &
    -\prescript{j}{i}{V_y}
\end{bmatrix}^{T} + \eta
\] where $\eta$ is the approximation error. This shows that using the aforementioned ``active'' control strategy, we obtain an implicit 3D structure of the environment in the optical flow. The inverse depth in this ``controlled'' case manifests as a linear function of the optical flow. 

The optical flow equation can be written for both foreground $\left(\mathcal{F}\right)$ and background $\left(\mathcal{B}\right)$ pixels independently. The magnitude of optical flow for $\mathcal{F}$ is given by,

\[\norm{\prescript{j}{i}{\dot{p}_{\mathbf{x}, \mathcal{F}}}}_2=
\prescript{j}{i}{Z_{\mathbf{x}, \mathcal{F}}}^{-1}
    \sqrt{\prescript{j}{i}{V_x^2} + \prescript{j}{i}{V_y^2}}
 + \nu
\]
where $\nu\sim\mathcal{N}(0,\sigma)$ is assumed to be an additive white Gaussian noise and is independent of the scene structure or the amount of camera movement between two frames ($V, \Omega$). For such assumptions to be valid, the optical flow algorithm needs to work well for a wide range of camera movements in a variety of scene structures. Using fast traditional optical flow formulations based on methods like \cite{LucasKanadeFlow} or \cite{HornSchunckFlow} voids such assumptions. This motivated us to use deep-learning based flow algorithms which excel at this task while maintaining a reasonable speed when running on a GPU. In this paper, FlowNet2 \cite{FlowNet2} is used to compute optical flow unless stated otherwise. 

A simple method to reduce noise is to compute the mean of the flow magnitudes across a few frames. Let $\xi = \{{\mathbb{F}^j, ..., \mathbb{F}^k}\}$ be a set of $N$ frames from which the optical flow is computed with respect to some reference frame $\mathbb{F}^i$ where the complete gap is assumed to be visible. Here, $N=\overline{\overline{\xi}}$ is the cardinality of the set $\xi$. The mean flow magnitude at $\mathbf{x}$ for $\mathcal{F}$ $\left(\norm{\prescript{\xi}{i}{\dot{p}_{\mathbf{x}, \mathcal{F}}}}_2\right)$ and $\mathcal{B}$ $\left(\norm{\prescript{\xi}{i}{\dot{p}_{\mathbf{x}, \mathcal{B}}}}_2\right)$ is given by,

\[\norm{\prescript{\xi}{i}{\dot{p}_{\mathbf{x}, \mathcal{F/B}}}}_2=
\left({N Z_{\mathbf{x},\mathcal{F/B}}}\right)^{-1}\sum_{r = j}^{k}{\prescript{r}{i}{V}} + \nu'\]

where $\nu'\sim\mathcal{N}\left(0,N^{-0.5}\sigma \right)$ and $V=\sqrt{V_x^2 + V_y^2}$. Clearly, the noise varies inversely with $N$.

Since $Z_{\mathbf{x},\mathcal{F}} < Z_{\mathbf{x},\mathcal{B}}$, we can say that $\norm{\prescript{\xi}{i}{\dot{p}_{\mathbf{x}, \mathcal{F}}}}_2 > \norm{\prescript{\xi}{i}{\dot{p}_{\mathbf{x}, \mathcal{B}}}}_2$.
Now, $\norm{\prescript{\xi}{i}{\dot{p}_{\mathbf{x}, \mathcal{F}}}}_2 - \norm{\prescript{\xi}{i}{\dot{p}_{\mathbf{x}, \mathcal{B}}}}_2 \geq \tau$ can be used as a criterion for finding possible boundary regions. It was found experimentally that using inverse flow differences $\norm{\prescript{\xi}{i}{\dot{p}_{\mathbf{x}, \mathcal{B}}}}_2^{-1} - \norm{\prescript{\xi}{i}{\dot{p}_{\mathbf{x}, \mathcal{F}}}}_2^{-1} \geq \tau'$ gave better numerical stability and better noise performance due to the scale compression by the inverse function. This is inherently the spatial derivative of inverse average (stacked) flow magnitudes and it can be written as
$\Xi = \nabla\cdot\norm{\prescript{\xi}{i}{\dot{p}_{\mathbf{x}}}}_2^{-1}$, where, $\nabla = \begin{bmatrix} \nicefrac{\partial}{\partial x} & \nicefrac{\partial}{\partial y}\end{bmatrix}^T$. Note that this is the same as solving the edge detection problem in computer vision and any kernel or method like the Sobel operator or the Canny edge detection algorithm can be used.

\section{High Speed Gap Tracking For Visual Servoing Based Control}
\label{sec:VisualServo}
This section presents a targeted solution for tracking a contour using label sets propagated using Focus of Expansion (FOE) constraints. 
A pixel at location $\mathbf{x}$ is associated with a score $\chi\left(\mathbf{x}\right) \in \left[-1,1 \right]$ which denotes its score as foreground or background. The foreground and background pixel locations are defined by $\mathcal{F} = \left\{ \mathbf{x} \vert \chi \left( \mathbf{x}\right) = +1 \right\}$ and $\mathcal{B} =\left\{ \mathbf{x} \vert \chi \left( \mathbf{x}\right) = -1 \right\} $ respectively.

\begin{figure}[t!]
    \centering
    \includegraphics[width=0.8\columnwidth]{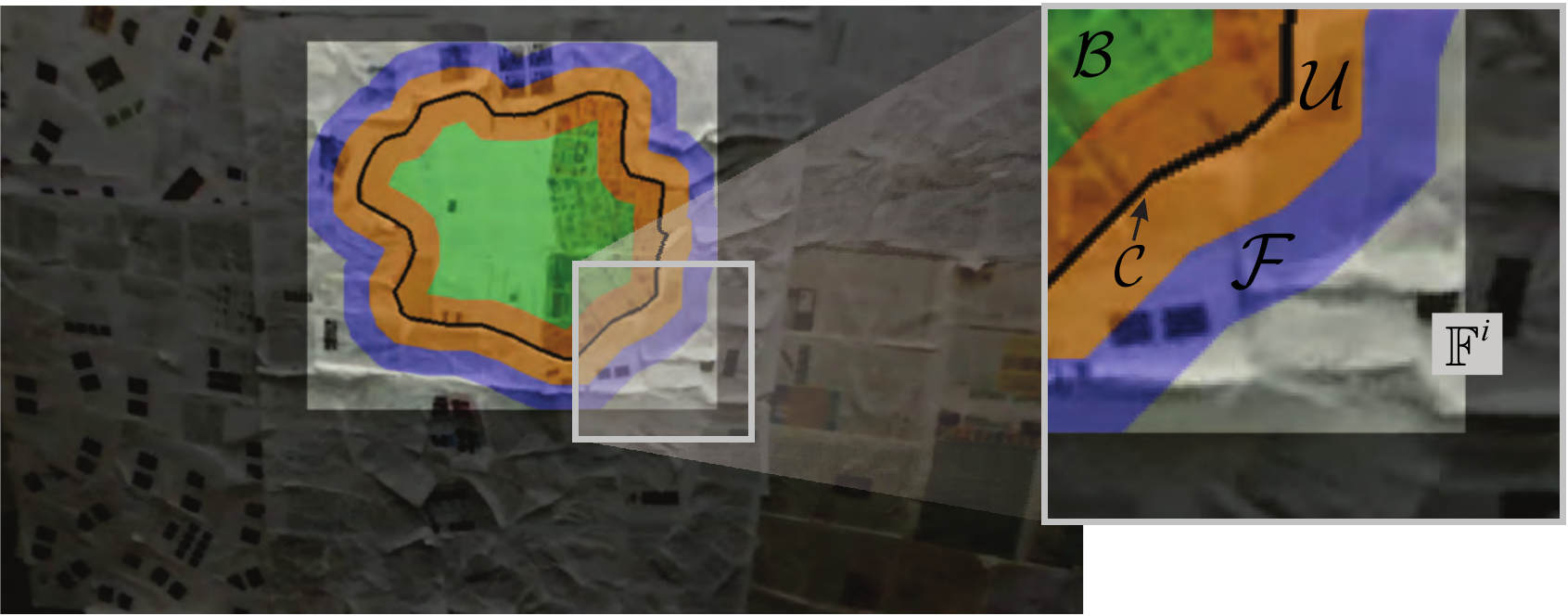}
    \caption{Label sets used in tracking. \textit{(blue: foreground region, green: background region, orange: uncertainty region, black line: contour, brighter part of frame: active region and darker part of frame: inactive region.)}}
    \label{fig:LabelSets}
\end{figure}

We define the opening $\mathcal{O}$ on the image plane as $\mathcal{O} = \left\{ \mathbf{x} \vert \chi \left( \mathbf{x} \right) < 0  \right\}$. The pixel locations which cannot be perfectly classified as foreground or background belong to the uncertainty zone and are defined as $\mathcal{U} =\left\{ \mathbf{x} \vert \chi \left( \mathbf{x}\right) \in \left(-1, +1\right) \right\} $. The contour location is defined as $\mathcal{C} =\left\{ \mathbf{x} \vert \chi \left( \mathbf{x}\right) = 0 \right\}$. 
Fig. \ref{fig:LabelSets} gives a visual representation of the different sets on a sample image.

The problem statement dictates the tracking of contour location $\mathcal{C}$ across time. This problem is hard as the contour tracking relies on updating $\chi\left( \mathbf{x} \right)\, \forall \, \mathbf{x} \in \mathcal{U}$ over time which is non-trivial and computationally expensive. To simplify the problem, we use the dual formulation of the problem which is to track the pixel locations which belong to the set defined by $\left\{ \mathbf{x} \notin \mathcal{U}\right\} = \mathcal{F} \cup \mathcal{B}$. This enables us to track the contour indirectly at high speeds as corner tracking is comparatively faster \cite{KLT}. The trade-off in using the dual formulation is that we don't obtain the actual contour location across time - which might be needed for aggressive maneuvers, but this is not dealt in the scope of this paper. The label set propagation algorithm is described in Algorithm \ref{algo:LabelSetProp}. Here, $\mathbb{C}_\mathcal{F}^i$ and $\mathbb{C}_\mathcal{B}^i$ represents a set of corner/feature points for the foreground and background pixels respectively in $\mathbb{F}^i$. $A^\dagger$ denotes the pseudo inverse of the matrix $A$ and $\begin{bmatrix} x_{0,\mathcal{F}} & y_{0,\mathcal{F}} \end{bmatrix}^T$ is the FOE for the foreground. Intuitively, we solve the linear equations of the horizontal flow field to obtain the divergence/time-to-contact $\alpha$. The divergence in x-direction is used to also predict the y-coordinates.

\begin{algorithm}[t!]
\KwData{$\mathbb{C}_{\mathcal{F}}^i, \mathbb{C}_{\mathcal{B}}^i, \mathbb{F}^i, \mathbb{F}^j, \mathcal{F}^i, \mathcal{B}^i$}
 \KwResult{$\mathbb{C}_{\mathcal{F}}^j, \mathbb{C}_{\mathcal{B}}^j, \mathcal{F}^j, \mathcal{B}^j$}
    $\mathbb{C}_\mathcal{F}^j$ = FeatureTracker$\left(\mathbb{C}_\mathcal{F}^i, \mathbb{F}^i, \mathbb{F}^j\right)$\;
    $A = \begin{bmatrix}
        \mathbb{C}_{\mathcal{F},x}^i & -\mathds{1}\\
    \end{bmatrix}$\;
    $B = \left[ \mathbb{C}_{\mathcal{F},x}^j - \mathbb{C}_{\mathcal{F},x}^i\right]$\;
    $\begin{bmatrix}
        \alpha\\ \beta
    \end{bmatrix} = A^\dagger B$\;
    $x_{0,\mathcal{F}} = \nicefrac{\beta}{\alpha};\qquad y_{0,\mathcal{F}} = \Big\langle \mathbb{C}_{\mathcal{F},y}^j  \dfrac{1}{\alpha} \left( \mathbb{C}_{\mathcal{F},y}^j - \mathbb{C}_{\mathcal{F},y}^i\right)\Big\rangle$\;
    $\mathcal{F}^j = \alpha \begin{bmatrix} \mathcal{F}^i_x - x_{0,\mathcal{F}} \\ \mathcal{F}^i_y - y_{0,\mathcal{F}} \end{bmatrix} + \begin{bmatrix} \mathcal{F}^i_x \\ \mathcal{F}^i_y\end{bmatrix} $ \;
    Repeat steps 1 through 6 for $\mathcal{B}$ in parallel\;
\caption{Label Set Propagation using FOE constraints.}
\label{algo:LabelSetProp}
\end{algorithm}

\subsection{Safe Point Computation and Tracking}
We can assume that the real-world points corresponding to the background $\mathcal{B} \cup \mathbb{U}_{B}$ are far enough that we can approximate them to lie on a plane. The foreground points under consideration $\mathcal{F} \cup \mathbb{U}_{F}$ occupy a small area around the contour which can be assumed to be planar. Here, $\mathbb{U}_{F} \subset \mathcal{U} \text{ and } \mathbb{U}_{B} \subset \mathcal{U}$ are the sets which actually belong to the foreground and background respectively.

Now, the quadrotor can be represented as an ellipsoid with semi-axes of lengths $a, b$ and $c$. As an implication of the above assumptions, the projection of the quadrotor on the window at any instant is an ellipse. Let us define the projection of the quadrotor on the image as $\mathcal{Q}$. The largest $\mathcal{Q}$ can be written in terms of matrix equation of the ellipse centered at $\left[ h, k\right]^T$ defined as $Q(h,k,R_\theta)=0$.

Here, $R_\theta$ is a two-dimensional rotation matrix and $\theta$ is the angle the largest semi-axis of the ellipse makes with the $X_C$ axis. The projection of the quadrotor on the image is given by $\mathcal{Q} = \left\{ \mathbf{x} \vert Q(\mathbf{x})\le 0\right\}$. The safe region $\mathcal{S}$ can be computed as \[\mathcal{S} = \bigcup_{\forall \theta} \mathcal{O}\ominus \mathcal{Q}\] where $\ominus$ denotes the Minkowski difference of sets. Now, we define the `safest point' ($\mathbf{x}_s$) as the barycentric mean of $\mathcal{S}$.

\begin{remark} The above optimization problem can only be solved using convex optimization with a guaranteed global solution when both $\mathcal{O}$ and $\mathcal{Q}$ are convex sets. A conservative solution to the above problem is fitting the largest scaled version of $\mathcal{Q}$ inside $\mathcal{O}$ when $\mathcal{O}$ is a non-convex set and $\mathcal{Q}$ is a convex set.
\end{remark}

Note that as $\mathcal{Q}$ is a chosen model, it can always be chosen to be a convex set, i.e., convex hull of the non-convex set. Also, from the above remark, the `safest point' ($\mathbf{x}_s$) can be defined as the center of the largest ellipse which can be fit inside $\mathcal{S}$ whilst maintaining the eccentricity equal to that defined by $\mathcal{Q}$. The optimization problem becomes,

\[ \argmax_{a,\theta} \mathcal{S} \cap \mathcal{Q} \text{ s.t. } \mathcal{Q} \subseteq \mathcal{S} \text{ and } \abs{\theta} \le \theta_{max}\]

This problem can be solved using the procedure described in \cite{Potato}. However, a desirable property of the safe region is that it should favor less-aggressive maneuvers. This can be modelled as a regularization penalty in the above formulation,
\[ \argmax_{a,\theta} \mathcal{S} \cap \mathcal{Q} + \lambda \theta \text{ s.t. } \mathcal{Q} \subseteq \mathcal{S} \text{ and } \abs{\theta} \le \theta_{max}\]

Solving the above optimization problem is computationally intensive and not-possible without the prior knowledge of the scale/depth. For obtaining the minimalist solution, we assume that \textit{the gap is large enough for the quadrotor to pass through} and replace the above optimization problem by an empirically chosen approximation. A simple and efficient safe point computation can be performed as the median of the convex set $\mathcal{O}$ and is given by $\mathbf{x}_s \approx \argmin_{\mathbf{x}} \sum_{\forall o \in \mathcal{O}} \Vert o - \mathbf{x}\Vert_2$. 

\begin{remark} If the above approximation is used when $\mathcal{O}$ is non-convex, the amount of deviation from the `actual' safe point is a function of $\sfrac{\overline{\overline{\text{Conv}\left( \mathcal{O}\right)}}}{\,\overline{\overline{\mathcal{O}}}}$. Here $\overline{\overline{\text{Conv}\left( \mathcal{O}\right)}}$ is the convex hull of $\mathcal{O}$.
\end{remark}

\begin{figure}[t!]
    \centering
    \includegraphics[width=\columnwidth]{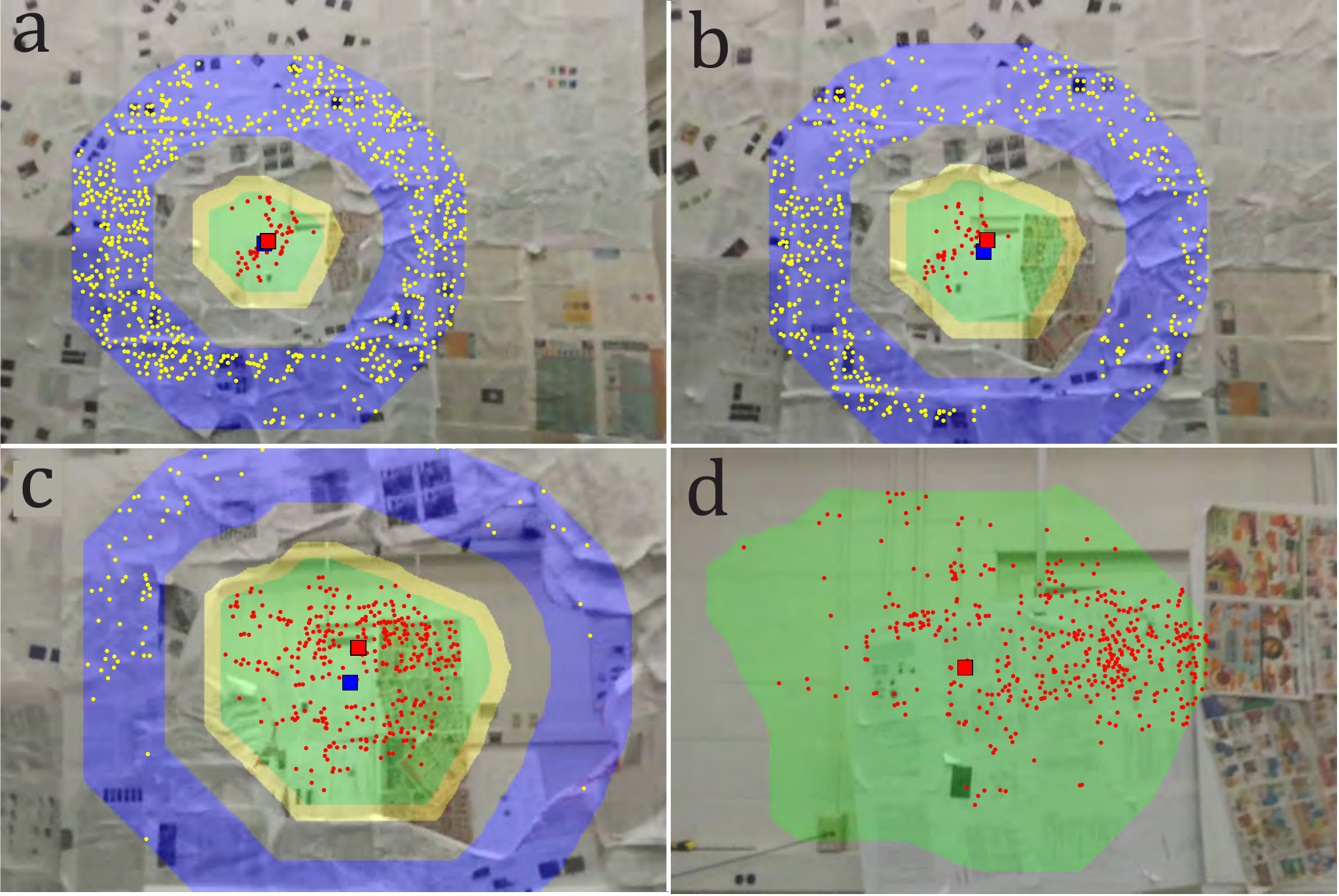}
    \caption{Tracking of $\mathcal{F}$ and $\mathcal{B}$ across frames. (a) shows tracking when $\overline{\overline{\mathbb{C}_{\mathcal{F}}^i}}> k_\mathcal{F}$ and $\overline{\overline{\mathbb{C}_{\mathcal{B}}^i}}> k_\mathcal{B}$. (b) When $\overline{\overline{\mathbb{C}_{\mathcal{B}}^i}}\le k_\mathcal{B}$, the tracking for $\mathcal{B}$ will be reset. (c) When $\overline{\overline{\mathbb{C}_{\mathcal{F}}^i}}\le k_\mathcal{F}$, the tracking for $\mathcal{F}$ will be reset. (d) shows tracking only with $\mathcal{B}$, when $\mathcal{F} =\emptyset$. \textit{(blue: $\mathcal{F}$, green:  $\mathcal{B}$, yellow: $\mathcal{O}$, yellow dots: $\mathbb{C}_{\mathcal{F}}^i$, red dots:  $\mathbb{C}_{\mathcal{B}}^i$, blue Square: $\mathbf{x}_{s,\mathcal{F}}$, red Square: $\mathbf{x}_{s,\mathcal{B}}$.)}}
    \label{fig:Tracking}
\end{figure}

Keen readers would note that the formulation for the safe point is in-terms of $\mathcal{O}$ which we wanted to avoid tracking in the first place. Indeed this is true and a simple solution takes care of this. Because we are propagating $\mathcal{F}$ and $\mathcal{B}$ with FOE constraints, the cross-ratios of $\left\{\norm{o,f}_2 \vert o\in\mathcal{O}, f\in\mathcal{F}\right\}$ and $\left\{\norm{o,b}_2 \vert o\in\mathcal{O}, b\in\mathcal{B}\right\}$ are preserved. Here, $\mathcal{F}$ and $\mathcal{B}$ are computed from the detected opening $\mathcal{O}$ as follows: $\mathcal{F} = \left\{\mathcal{O} \oplus \epsilon_1 - \mathcal{O} \oplus \epsilon_2\right\}$ and  $\mathcal{B} = \left\{\mathcal{O} \ominus \epsilon_3\right\}$. Here, $\epsilon_i$ is a user chosen kernel (circular in this paper).

The $\mathcal{F}$ and $\mathcal{B}$ are propagated from $\mathbb{F}^i$ onwards where the detection was performed. The `safest point' ($\mathbf{x}_s$) is computed as follows:
\[
\mathbf{x}_s =
     \begin{cases}

     \mathbf{x}_{s,\mathcal{F}} , &\quad \overline{\overline{\mathcal{F}}}\geq\overline{\overline{\mathcal{B}}} \\

       \mathbf{x}_{s,\mathcal{B}}, &\quad \text{otherwise}\\
     \end{cases}
\]
where $\mathbf{x}_{s, \mathcal{F}}$ and $\mathbf{x}_{s, \mathcal{B}}$ are the safest points computed individually for $\mathcal{F}$ and $\mathcal{B}$ by taking their median respectively. Fig. \ref{fig:Tracking} shows the tracking of $\mathcal{F}$ and $\mathcal{B}$ across time and how the tracking algorithm actively switches between $\mathbf{x}_{s,\mathcal{F}}$ and $\mathbf{x}_{s,\mathcal{B}}$ for the safest point $\mathbf{x}_s$. When  $\overline{\overline{\mathbb{C}_{\mathcal{F}}^i}}\le k_\mathcal{F}$ or $\overline{\overline{\mathbb{C}_{\mathcal{B}}^i}}\le k_\mathcal{B}$, the tracker/feature points are reset in the current $\mathcal{F}$ and $\mathcal{B}$ sets respectively. Here, $k_\mathcal{F}$ and $k_\mathcal{B}$ are empirically chosen thresholds. For all experiments $k_\mathcal{F}=40$ and $k_\mathcal{B}=20$. Due to resetting and active switching, $\mathbf{x}_{s}$ can jump around making the control hard, hence a simple Kalman filter with a forward motion model is used to smooth out the value of $\mathbf{x}_{s}$. From here on, safe point refers to the safest point $\mathbf{x}_{s}$.

\subsection{Control Policy}
We propose a control policy such that it follows the tracked $\mathbf{x}_s$. The quadrotor follows the dynamic model as given in \cite{MinimumSnapTrajectory}. The controller uses the traditional backstepping approach based on \cite{MinimumSnapTrajectory} and contains the following loops: Inner loop and outer loop controllers. Inner loop controls the attitude stability while the outer loop controller is responsible for the quadrotor position. It is important to note that frames are transformed from $C$ to $B$.

Since the quadrotor is differentially flat, the altitude $Z_B$ can be controlled independently from $X_B$ and $Y_B$ \cite{DifferentialFlatness}. The control policy is to align the projection of the body center on the image plane with $\mathbf{x}_s$. The difference between the two centers is called the error $e$. The $x$ and $y$ component of the error $e$ can be minimized by varying roll $\left(\phi\right)$ and net thrust $\left(u_1\right)$ respectively. A simple Proportional-Integral-Derivative (PID) controller on $e$ is used. This control policy only deals with the alignment of $B$ to $\mathbf{x}_s$ and does not deal with moving forward ($Z_C$). To move forward, the quadrotor pitch $\left(\phi\right)$ needs to be controlled. The rate of forward motion is controlled by the pitch angle $\theta_0$ which is empirically chosen. The bigger the value of $\theta_0$ the faster the quadrotor will fly towards the gap. It is important to note the implicit assumption made in this paper that the gap is large enough for the quadrotor to go through safely.

\section{Experiments}
\label{sec:Expts}
\subsection{Experimental Setup}
\begin{figure}[t!]
    \centering
    \includegraphics[width=0.7\columnwidth]{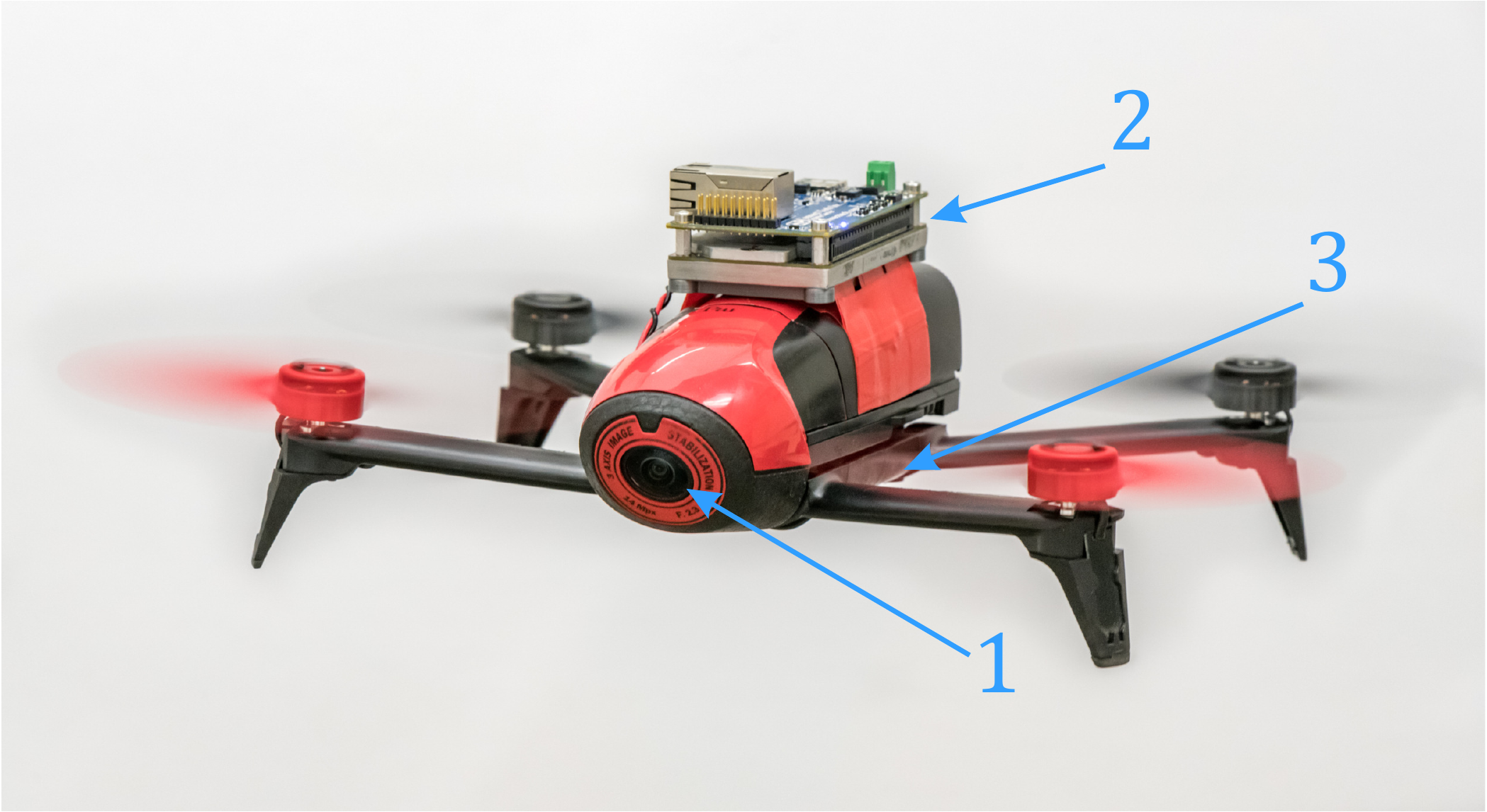}
    \caption{The platform used for experiments. (1) The front facing camera, (2) NVIDIA TX2 CPU+GPU, (3) Downward facing optical flow sensor (camera+sonar) which is only used for position hold.}
    \label{fig:QuadrotorParts}
\end{figure}

\begin{figure}[t!]
    \centering
    \includegraphics[width=\columnwidth]{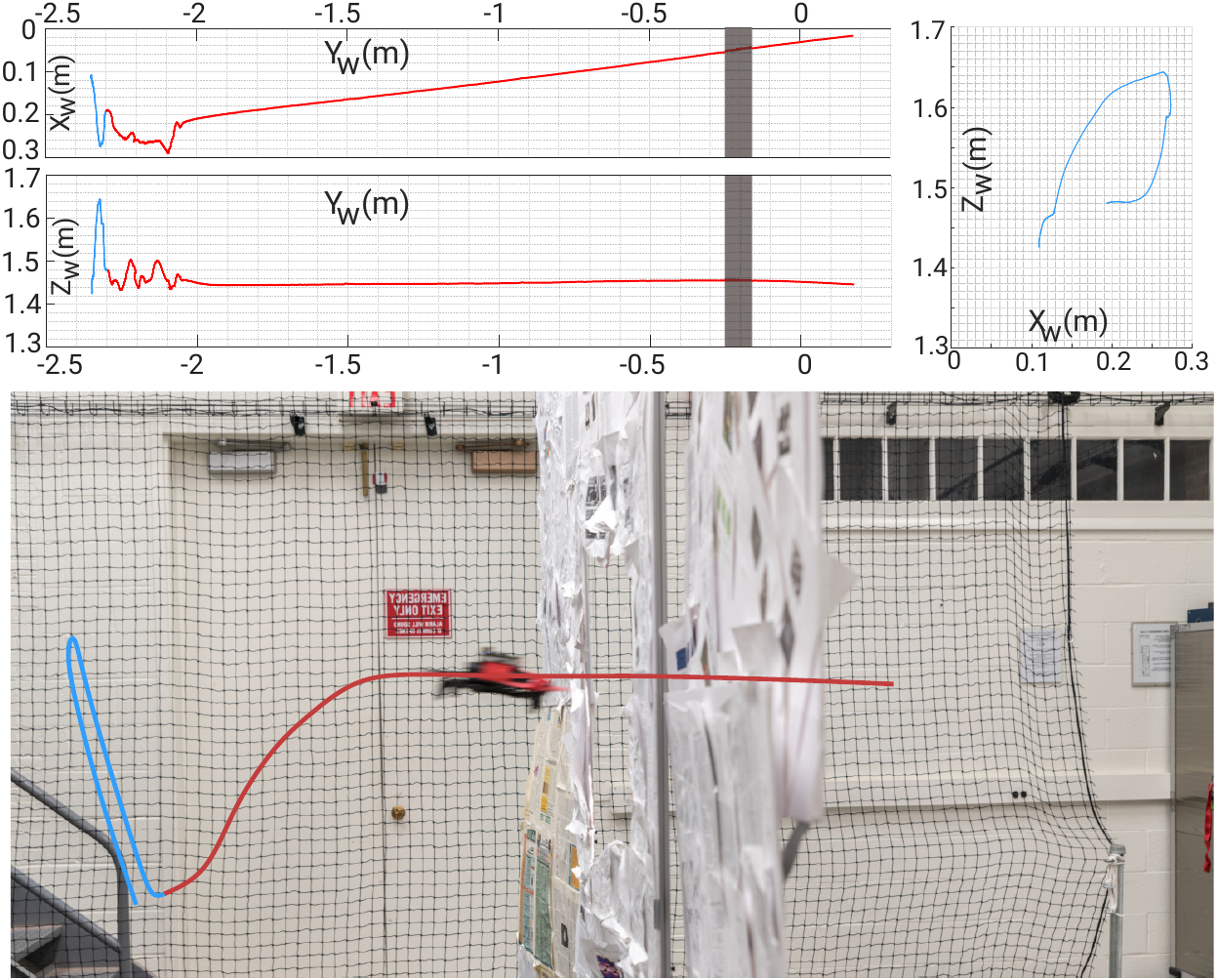}
    \caption{First two rows: $\left(X_W,Y_W\right)$, $\left(Y_W,Z_W\right)$ and $\left(X_W,Z_W\right)$ Vicon estimates of the trajectory executed by the quadrotor in different stages (gray bar indicates the gap). $\left(X_W,Z_W\right)$ plot shows the diagonal scanning trajectory (the lines don't coincide due to drift). Last row: Photo of the quadrotor during gap traversal. \textit{(cyan: detection stage, red: traversal stage.)}}
    \label{fig:QuadrotorDetectionAndServo}
\end{figure}

The proposed framework was tested on a modified hobby quadrotor, Parrot$^\text{\textregistered}$  Bebop 2, for its cost effectiveness and ease of use. The Bebop 2 is equipped with a front facing camera, a 9-axis IMU and a downward facing optical flow sensor coupled with a sonar.
The Parrot$^\text{\textregistered}$ Bebop 2 allows only high level controls in terms of body frame velocities using ROS. An NVIDIA Jetson TX2 GPU is mounted on the Bebop 2 as shown in Fig. \ref{fig:QuadrotorParts} and is used to run all the perception and control algorithms onboard. The TX2 and Bebop 2 communicate via a WiFi connection, where the images are received at 30Hz. The overall weight of the flight setup is $680$g with the dimensions being $32.8 \times 38.2 \times 12$cm.\\
\indent All the experiments were prototyped on a PC running Ubuntu 16.04 with an Intel$^\text{\textregistered}$ Core i7 6850K 3.6GHz CPU, an NVIDIA Titan-Xp GPU and 64GB of RAM in \textsc{Matlab} using the Robotics Toolbox. The deep learning based optical flow runs on Python with TensorFlow back-end. All the final versions of the software were ported to Python to run on the NVIDIA Jetson TX2 running Linux for Tegra$^\text{\textregistered}$ (L4T) 28.2. A step-by-step tutorial on using Bebop 2 as a research platform is available at \url{prg.cs.umd.edu/GapFlyt.html}.\\
\indent The environmental setup for the experiments consists of a rigid scene which has two near-planar structures, one for the foreground and the other for the background. As shown in Fig. \ref{fig:FgBgDist}, let us denote the initial perpendicular distance between the quadrotor body center and the foreground as $\prescript{0}{}{Z_{\mathcal{F}}}$ and the background as $\prescript{0}{}{Z_{\mathcal{B}}}$. The near-planar structures are made of foam-core with newspapers stuck on them to add texture to the scene. The gap is located near the center of the foreground and is of an arbitrary shape. For the detection of the window, the quadrotor executes a fixed diagonal straight line trajectory in the $X_W-Z_W$ plane as shown in Fig. \ref{fig:QuadrotorDetectionAndServo} while taking a number of images along its path. The number of images used for the detection stage is a parameter denoted by $N$. Once the window is detected, $\mathcal{F}$ and $\mathcal{B}$ are tracked across time in order for quadrotor to go through the gap using visual servoing based control.

\begin{figure*}[t!]
    \centering
    \includegraphics[width=\textwidth]{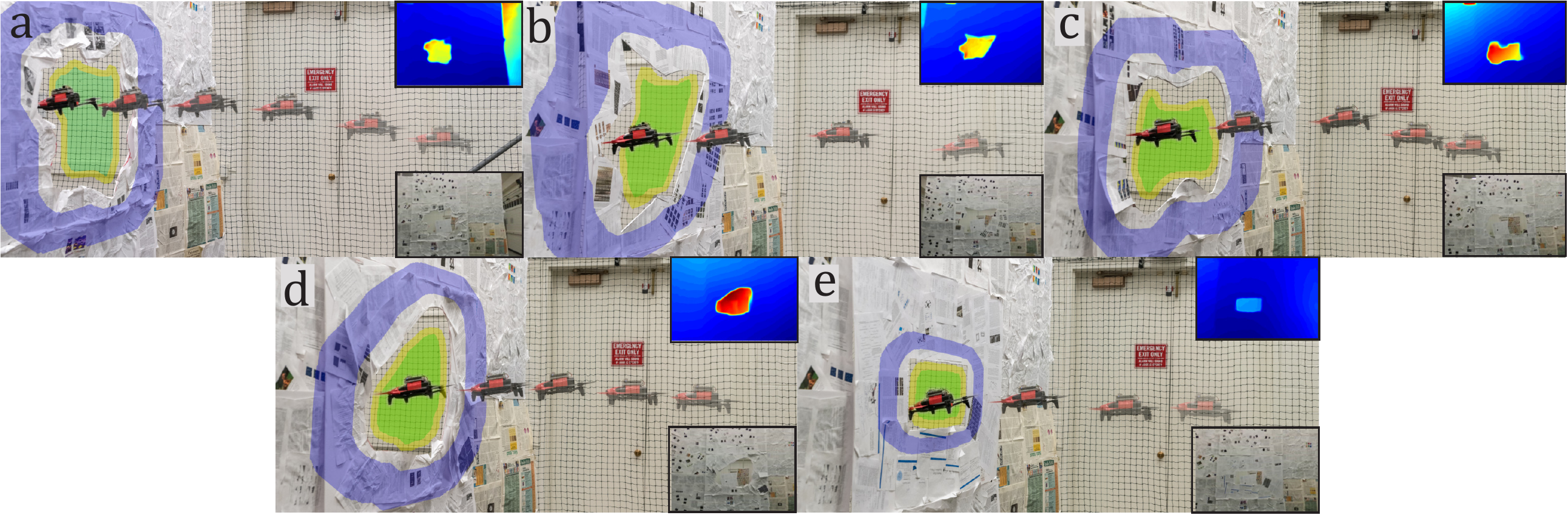}
    \caption{Sequence of images of quadrotor going through different shaped gaps. Top on-set: $\Xi$ outputs, bottom on-set: quadrotor view.}
    \label{fig:AllWindows}
\end{figure*}
\subsection{Experimental Results}
The pipeline was evaluated on different variations of the environmental setup. In the first experiment, we test our pipeline on five different arbitrarily shaped gaps as shown in Fig. \ref{fig:AllWindows}. Unless otherwise stated $\prescript{0}{}{Z_{\mathcal{F}}} \sim 2.6$m and  $\prescript{0}{}{Z_{\mathcal{B}}} \sim 5.7$m. The aim here is to find biases in the detection pipeline. The windows were chosen to have a diversity in the geometrical sharpness of the corners, convexity of the shape and the size. As stated earlier, the only constraint imposed on the gaps is that they are large enough to go through with the quadrotor pitch angle close to zero and near-convex. The outputs of TS$^2$P algorithm for different windows are shown in Fig. \ref{fig:AllWindows} with $N=4$ along with their inverse  average (stacked)  flow magnitudes $\Xi$, illustrating that our detection algorithm is independent of shape and size of the opening. A canny edge detector is run on $\Xi$ followed by morphological operations to obtain $\mathcal{C}$.

\begin{figure}[t!]
    \centering
    \includegraphics[width=\columnwidth]{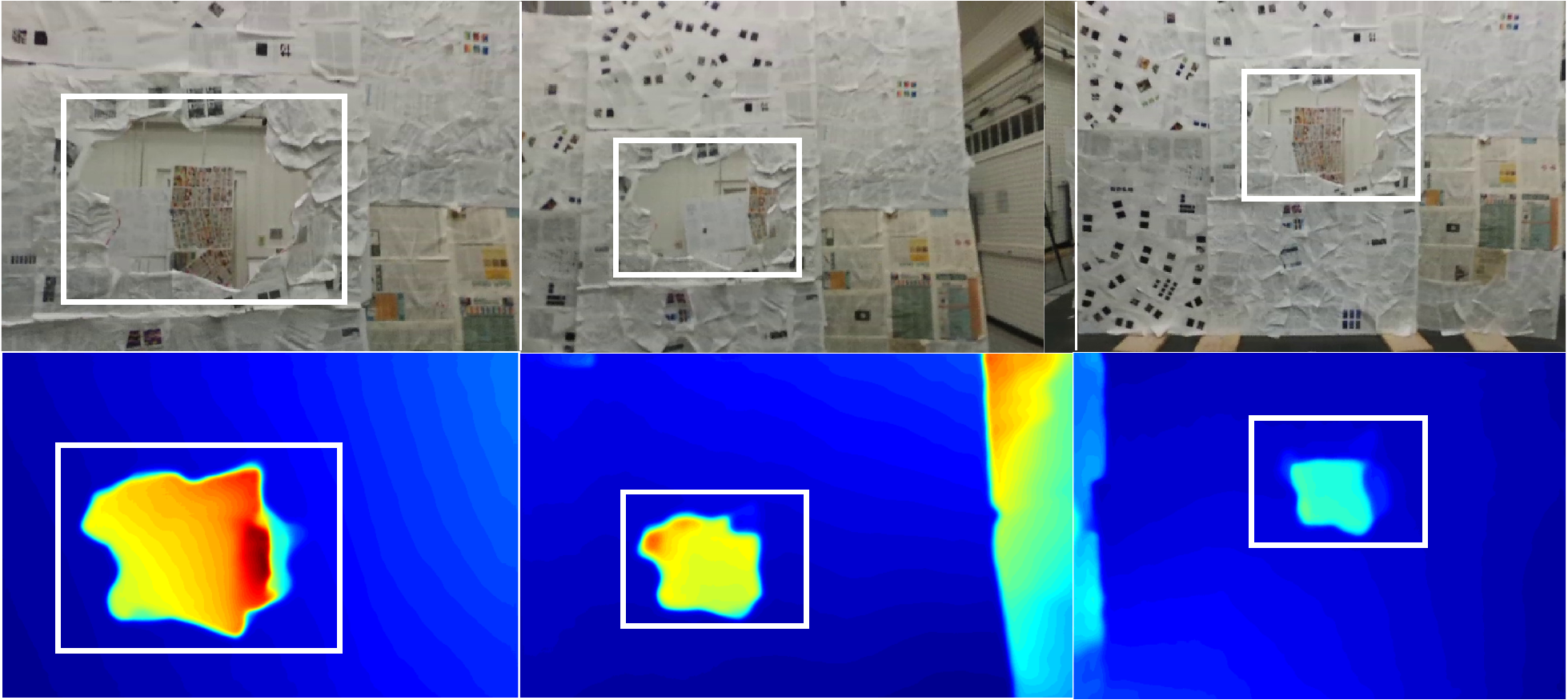}
    \caption{Top Row (left to right): Quadrotor view at $\prescript{0}{}{Z_{\mathcal{F}}} = 1.5, 2.6, 3$m respectively with $\prescript{0}{}{Z_{\mathcal{B}}}=5.7$m. Bottom Row: Respective $\Xi$ outputs for $N=4$. Observe how the fidelity of $\Xi$ reduces as $\prescript{0}{}{Z_{\mathcal{F}}} \to \prescript{0}{}{Z_{\mathcal{B}}}$, making the detection more noisy. \textit{(white boxes show the location of the gap in Figs. \ref{fig:DiffDist} to \ref{fig:DiffN}.)}}
    \label{fig:DiffDist}
\end{figure}

The second experiment is designed to test the noise sensitivity of TS$^2$P. The intuition is that as $\prescript{0}{}{Z_{\mathcal{F}}}\to\prescript{0}{}{Z_{\mathcal{B}}}$, noisier the detection result . The outputs for different $\prescript{0}{}{Z_{\mathcal{F}}}$ and $\prescript{0}{}{Z_{\mathcal{B}}}$ are shown in Fig. \ref{fig:DiffDist} when $N=4$. This is because the fidelity of $\Xi$ becomes less and is more prone to noise. By increasing $N$ the noise gets averaged out across frames improving the fidelity of $\Xi$.

\begin{figure}[t!]
    \centering
    \includegraphics[width=\columnwidth]{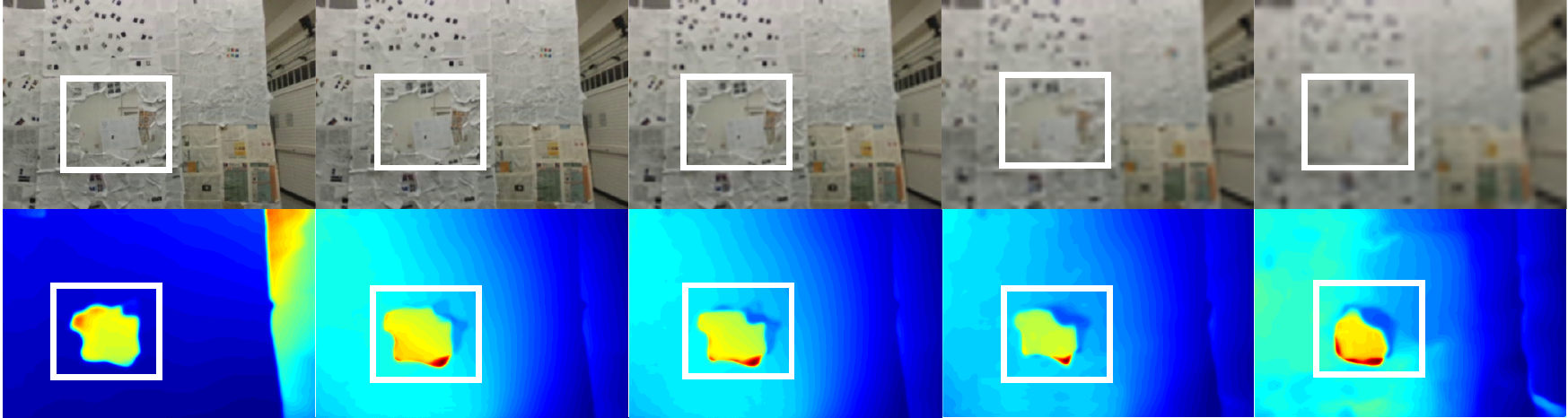}
    \caption{Top row (left to right): Quadrotor view at image sizes of $384 \times 576$, $192 \times 288$, $96 \times 144$, $48 \times 72$, $32 \times 48$. Note all images are re-scaled to $384 \times 576$ for better viewing. Bottom row shows the respective $\Xi$ outputs for $N=4$.}
    \label{fig:DiffImageSizes}
\end{figure}

\begin{figure}[t!]
    \centering
    \includegraphics[width=0.9\columnwidth]{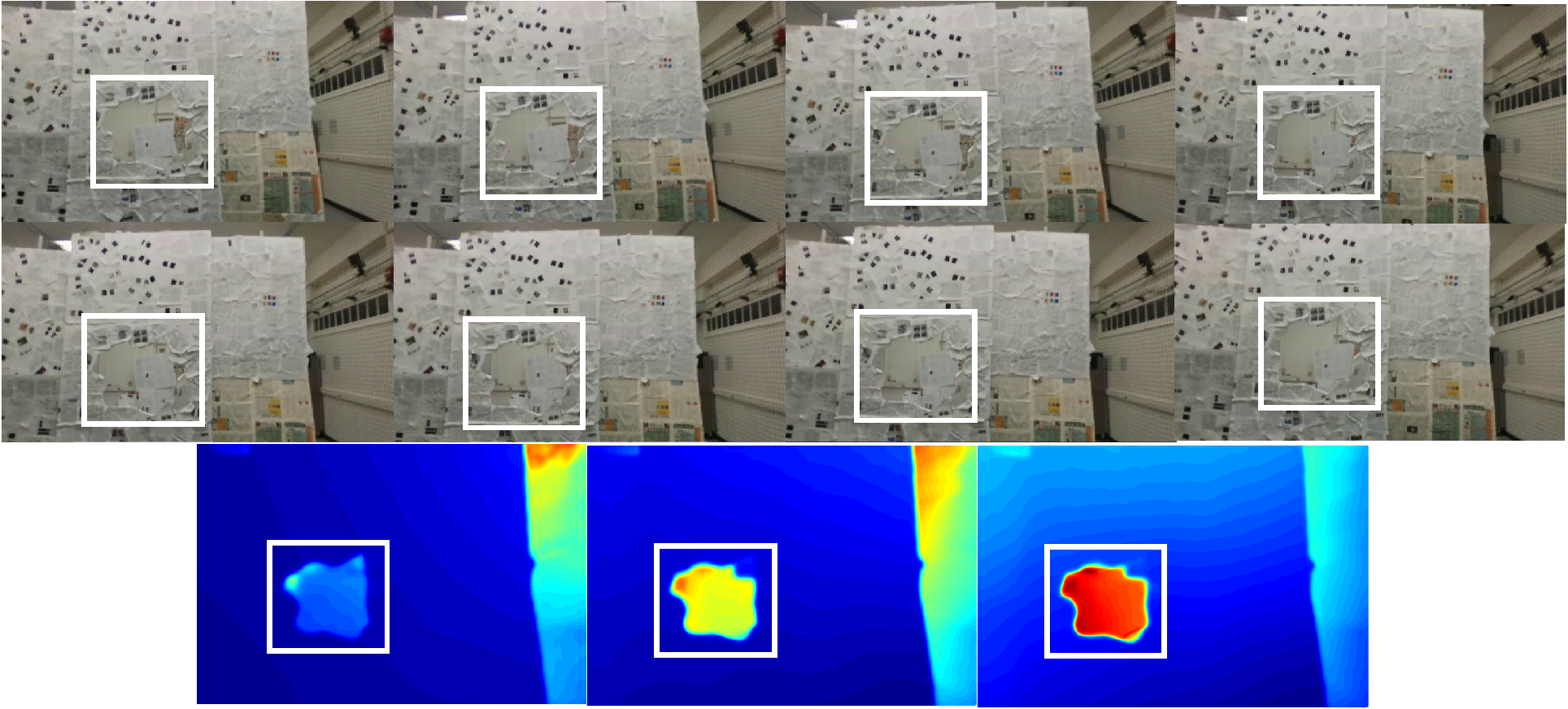}
    \caption{Top two rows show the input images. The third row shows the $\Xi$ outputs when only the first 2, 4 and all 8 images are used.}
    \label{fig:DiffN}
\end{figure}

In the third experiment, we present detection outputs for different values of $N$, image baselines and image sizes. The effect of $N$ has been already discussed previously. Having a very small baseline results in effectively dropping the value of $N$ and vice-versa. The results from different sized images as illustrated in Fig. \ref{fig:DiffImageSizes} show that the detection algorithm can work even on a very small quadrotor which can only carry a very low-resolution camera (as low as 32 $\times$ 48 pixels). Our algorithm can also handle dynamic noises very well though being modelled as a gaussian for the discussion. However, one can notice that the results improve significantly with increase in $N$ (Fig. \ref{fig:DiffN}) demonstrating the advantage of TS$^2$P.

Gap detection using TS$^2$P almost always results in an eroded version of the true gap. This is good for safe maneuvers like the one considered in this paper. However, aggressive flight algorithms might suffer due to conservative results. This can be mitigated by tuning the values of image baselines, $N$ and the trajectory of the quadrotor to obtain minimum erosion results. Tuning these parameters is easy when a prior about the scene is known or the flow algorithms are so fast that one can actively change the detection trajectory so as to maximize the coverage on the part of the contour least `seen'. The dynamic choice of these parameters comes into the scope of our future work.

In the last experiment we present alternative approaches including state-of-the-art methods which can be used to find the gap. The methods can be subdivided into structure based approaches and stuctureless approaches. The structure based approaches can be defined as the set of approaches where a full 3D reconstruction of the scene is computed, whereas, stuctureless approaches do not. The structure based approaches presented are DSO \cite{DSO} -- Direct Sparse Odometry, depth from hardware stereo cameras \cite{SLAMDunk} and Stereo SLAM -- Simultaneous Localization and mapping using Stereo Cameras and IMU \cite{SLAMDunk}. The data for the structured approaches were collected using a Parrot$^\text{\textregistered}$ SLAMDunk \cite{SLAMDunk}. The structureless approaches presented are MonoDepth \cite{MonoDepth} -- deep learning based monocular depth estimation and the proposed TS$^2$P on two different deep learning based dense optical flow algorithms, namely, FlowNet2 \cite{FlowNet2}, SpyNet \cite{SpyNet} and DIS \cite{DIS}. Table \ref{tab:DetectionTable} shows the comparison of the stated methods averaged over 150 trials. 

\begin{figure}[t!]
    \centering
    \includegraphics[width=\columnwidth]{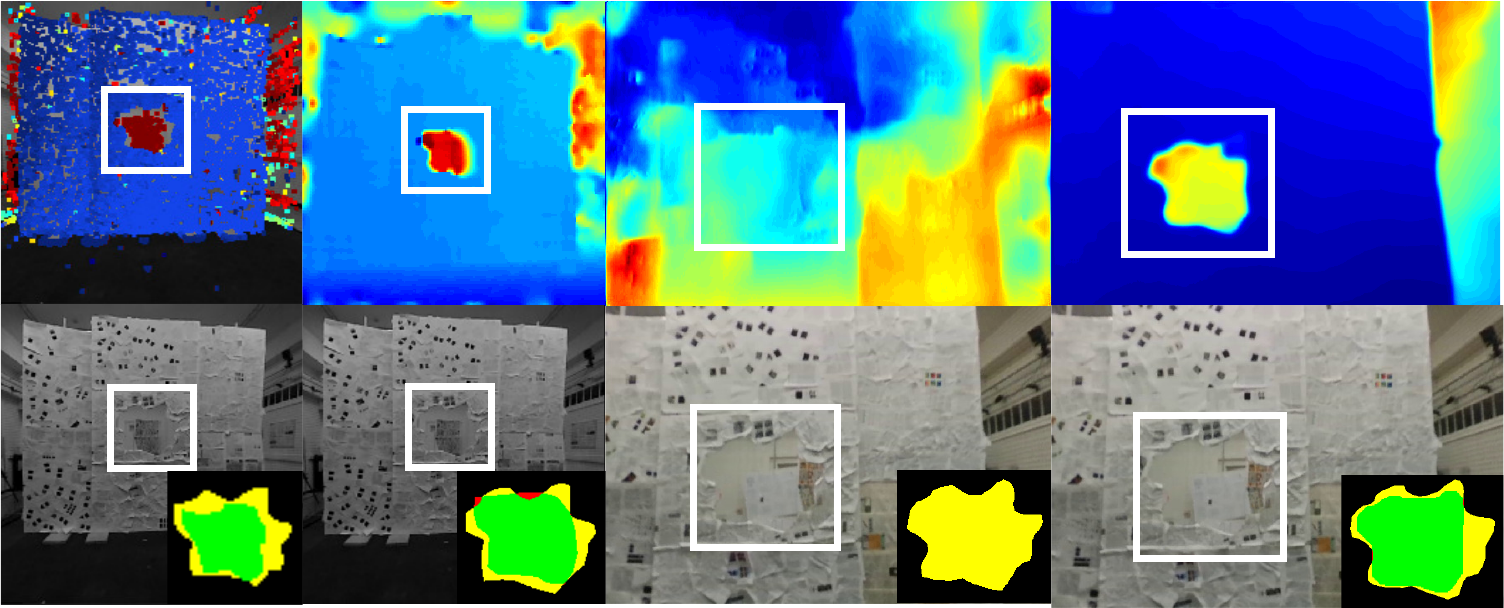}
    \caption{Comparison of different philosophies to gap detection. Top row (left to right): DSO, Stereo Depth, MonoDepth, TS$^2$P. Bottom row shows the detected gap overlayed on the corresponding input image. \textit{(green: $\mathcal{G} \cap \mathcal{O}$, yellow: false negative $\mathcal{G} \cap \mathcal{O}^{\prime}$, red: false positive $\mathcal{G}^{\prime} \cap \mathcal{O}$.)}}
    \label{fig:StructuredVsUnstructuredApproaches}
\end{figure}

Fig. \ref{fig:StructuredVsUnstructuredApproaches} compares the results of DSO, stereo depth, MonoDepth and our method (TS$^2$P) with the ground truth. It can be inferred that the MonoDepth results are extremely noisy (even with different models) making it impossible to detect the gap as the images in our paper were never ``seen'' during training. Note that we don't retrain or finetune any of the deep learning models in this paper. Retraining MonoDepth and other deep learning based methods used in this paper on our dataset might lead to better results. Whereas, DSO and stereo depth results can used to detect the opening with some filtering. Stereo SLAM and DSO are slow in the map building stage (taking about 6s and 12s respectively), however, once the map is built and the algorithms are in the localization stage the depth (or scaled depth) are obtained at 20Hz. The Stereo SLAM and Stereo Depth were run on the SLAMDunk with an NVIDIA Jetson TK1 processor which is much slower than the NVIDIA Jetson TX2 processor used for running DSO and other methods. 

Fig. \ref{fig:DiffFlowAlgorithms} compares different optical flow methods used for TS$^2$P. Though SpyNet and DIS optical flow are faster, FlowNet2 outputs significantly better results at the edges which is important for obtaining a good gap detection -- this can be observed by looking at $\Xi$ for each algorithm.

After the gap detection has been performed, $\mathcal{F}$ and $\mathcal{B}$ are computed from the detected gap $\mathcal{C}$. Fig. \ref{fig:Tracking}  shows $\mathcal{F}$ and $\mathcal{B}$ being propagated across time as the quadrotor is in pursuit of going through the gap with the update of $\mathbf{x}_s$. A comparison of tracking using different methods are given in Table \ref{tab:TrackingTable}. Clearly, KLT outperforms all other methods with a theoretical maximum quadrotor speed of 8 ms$^{-1}$ in the given scene. The theoretical maximum speed is calculated for a global shutter camera in such a way that the motion parallax is constrained within one pixel for the scene with  $\prescript{0}{}{Z_{\mathcal{F}}} \sim 2.6$m and  $\prescript{0}{}{Z_{\mathcal{B}}} \sim 5.7$m. The calculation assumes that none of the tracking/matching methods work when the motion blur is more than one pixel. However, most of the methods can work well upto some pixels of motion blur, this will in-turn increase the theoretical maximum speed by this factor. If a rolling shutter camera is used without rolling shutter compensation, the theoretical maximum speed value has to be divided by the factor of blur caused by rolling shutter. We achieved a practical maximum speed of 2.5ms$^{-1}$ in our experiments. We were limited to this speed due to the acceleration constraints on the Bebop 2 and the rolling shutter camera.

\begin{figure}[t!]
    \centering
    \includegraphics[width=0.7\columnwidth]{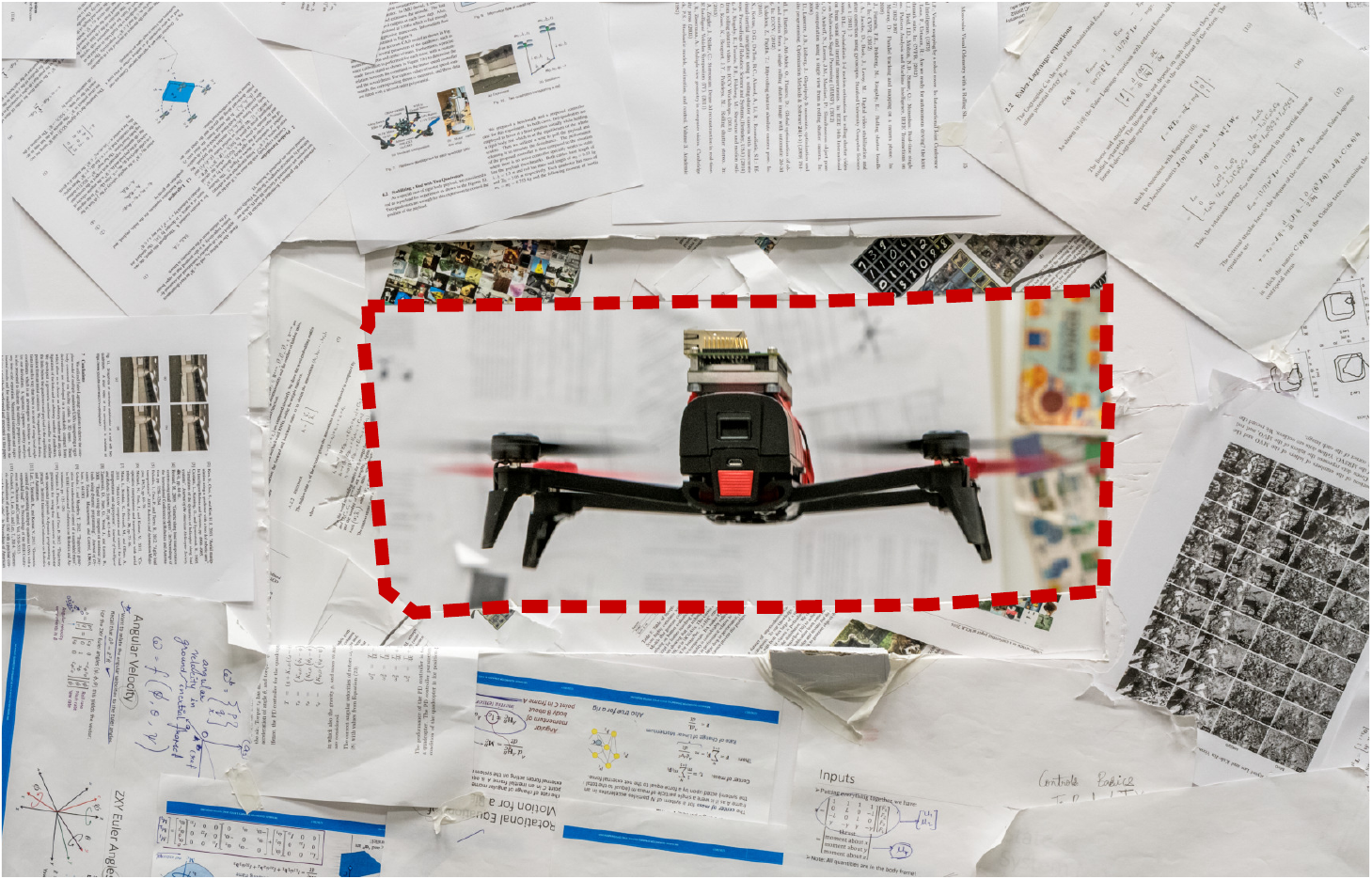}
    \caption{Quadrotor traversing an unknown window with a minimum tolerance of just 5cm. \textit{(red dashed line denotes $\mathcal{C}$.)}}
    \label{fig:QuadRectWindow}
\end{figure}

We achieved a remarkable success rate of 85$\%$ over 150 trials for different arbitrary shaped windows under a wide range of conditions which includes a window with a minimum tolerance of just 5cm (Fig. \ref{fig:QuadRectWindow}). Success is defined as window detection output $\mathcal{O}$ having at least 75\% overlap with the ground truth and traversal through the gap without collision. Failure cases also include the optimization failures and/or feature tracking failures for structure based approaches. For TS$^2$P, we define Detection Rate (DR), Average False Negative (AFN) and Average False Positive (AFP) as follows (AFN and AFP are computed only for successful trails):

\begin{figure}[t!]
    \centering
    \includegraphics[width=\columnwidth]{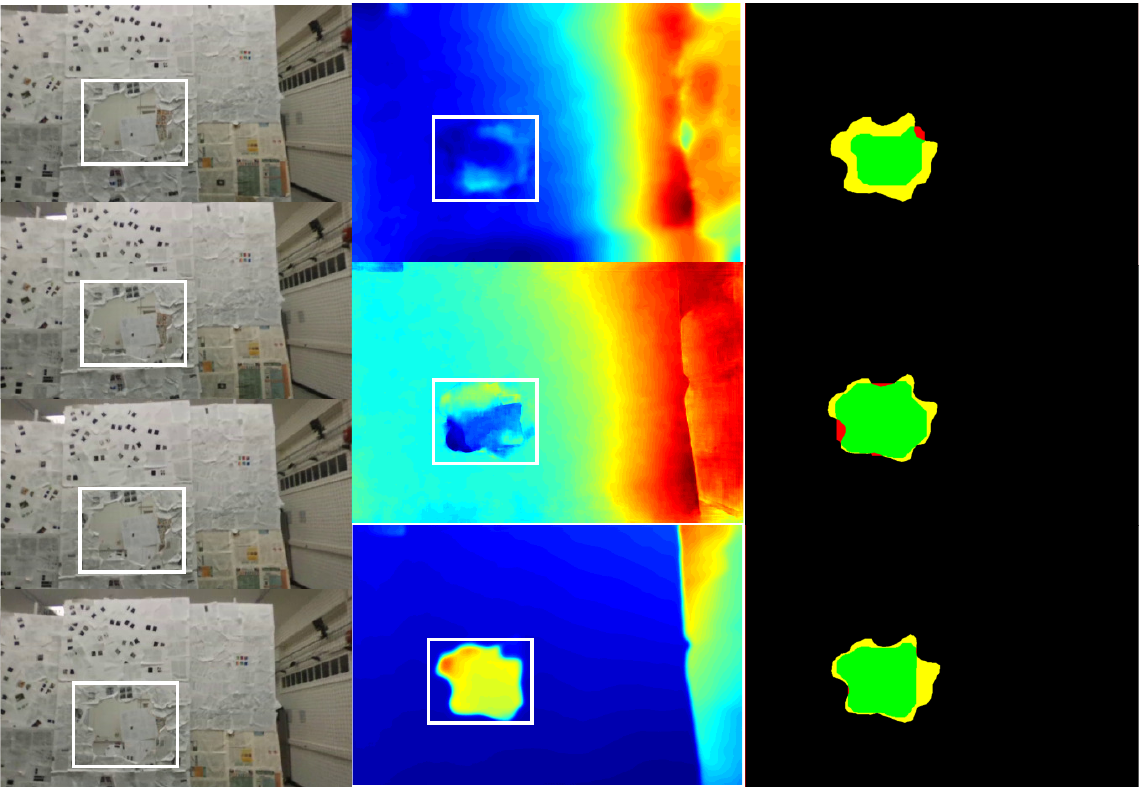}
    \caption{Left Column: Images used to compute $\Xi$. Middle Column (top to bottom): $\Xi$ outputs for DIS Flow, SpyNet and FlowNet2. Right Column: Gap Detection outputs. \textit{(green: $\mathcal{G} \cap \mathcal{O}$, yellow: false negative $\mathcal{G} \cap \mathcal{O}^{\prime}$, red: false positive $\mathcal{G}^{\prime} \cap \mathcal{O}$.}}
    \label{fig:DiffFlowAlgorithms}
\end{figure}

\begin{table}[t!]
\centering
\caption{Comparison of different methods used for gap detection.}
\resizebox{0.95\columnwidth}{!}{
\label{tab:DetectionTable}
\begin{tabular}{llllll}
\toprule
Method & Sensor(s) & Run Time & DR & AFN & AFP \\ & & (Init. Time) in s & & & \\
 \hline
 DSO\cite{DSO} & Monocular & 0.05 (6) & 0.88 & 0.20 & \textbf{0.00}\\
 Stereo Depth$^*$\cite{SLAMDunk} & Stereo & 0.10 & 0.90 & 0.17 & 0.04  \\
 Stereo SLAM$^*$\cite{SLAMDunk} & Stereo + IMU & 0.05 (12) & 0.91 & 0.15 & \textbf{0.00}\\
 Mono Depth\cite{MonoDepth} & Monocular & 0.97 & 0.00 & -- & --\\
 TS$^2$P (FlowNet2\cite{FlowNet2}) & Monocular & 1.00 & \textbf{0.93} & \textbf{0.14} & 0.02\\
 TS$^2$P (SpyNet\cite{SpyNet}) & Monocular &  0.12 & 0.74 & 0.16 & 0.05 \\
 TS$^2$P (DIS\cite{DIS}) & Monocular & 0.45 & 0.62 & 0.20 & 0.04 \\
 \bottomrule
\end{tabular}}
  \\\tiny{$^*$ indicates algorithm tested on NVIDIA TK1 otherwise NVIDIA TX2.}
\end{table}


\begin{equation*}
    \resizebox{0.75\columnwidth}{!}{$\text{DR} = \cfrac{\sum_{k=1}^{\text{Num. Trails}}\left(\lambda_D^k\right)}{\text{Num. Trails}}; \lambda_D^k = \left(\cfrac{\overline{\overline{\mathcal{G}\cap \mathcal{O}}}}{{\overline{\overline{\mathcal{G}}}}}\right)^k \geq 0.75$}
\end{equation*}

\begin{equation*}
    \resizebox{0.65\columnwidth}{!}{$\text{AFN} = \cfrac{\sum_{k=1}^{\text{Num. Succ. Trails}}\left(\lambda_N^k\right)}{\text{Num. Succ. Trails}}; \lambda_N^k = \left(\cfrac{\overline{\overline{\mathcal{G}\cap \mathcal{O^{\prime}}}}}{{\overline{\overline{\mathcal{G}}}}}\right)^k$}
\end{equation*}

\begin{equation*}
    \resizebox{0.65\columnwidth}{!}{$\text{AFP} = \cfrac{\sum_{k=1}^{\text{Num. Succ. Trails}}\left(\lambda_P^k\right)}{\text{Num. Succ. Trails}}; \lambda_P^k = \left(\cfrac{\overline{\overline{\mathcal{G^{\prime}}\cap \mathcal{O}}}}{{\overline{\overline{\mathcal{G}}}}}\right)^k$}
\end{equation*}

where $\mathcal{A}^\prime$ is the negation of the set $\mathcal{A}$.

\begin{table}[t!]
\centering
\caption{Comparison of different methods used for tracking.}
\resizebox{0.85\columnwidth}{!}{
\label{tab:TrackingTable}
\begin{tabular}{lll}
\toprule
Method & Run Time (ms) & Theo. Max. Speed (ms$^{-1}$) \\
 \hline
 GMS\cite{GMS} & 40 & 0.40 \\
 FAST\cite{FAST} + RANSAC & 8.3 & 1.92 \\
 Cuda-SIFT\cite{CUDASIFT} + RANSAC & 5 & 3.20  \\
 KLT\cite{KLT} & 2 & 8.00 \\
 \bottomrule
\end{tabular}}
\end{table}

\section{Conclusions}
\label{sec:Conc}
We present a minimalist philosophy to mimic insect behaviour to solve complex problems with minimal sensing and active movement to simplify the problem in hand. This philosophy was used to develop a method to find an unknown gap and fly through it using only a monocular camera and onboard sensing. A comprehensive comparison and analysis is provided. To our knowledge, this is the first paper which addresses the problem of gap detection of an unknown shape and location with a monocular camera and onboard sensing. As a parting thought, IMU data can be coupled with the monocular camera to get a scale of the window and plan for aggressive maneuvers.




\section*{Acknowledgement}
The authors would like to thank Konstantinos Zampogiannis for helpful discussions and
feedback.



\bibliographystyle{unsrt}
\bibliography{Ref}

\clearpage
\appendix
\subsection{Robustness of TS$^2$P against different textures}
Optical flow algorithms are generally biased to specific textures, and there is a high correlation between highly textured surfaces and good optical flow computation. To demonstrate the robustness of our approach (TS$^2$P) we test the algorithm against ten additional setups. The various scenarios are tabulated in Table \ref{tab:CasesTable}. Each scenario is a combination of different textures from the following set: Bumpy, Leaves, Door, Newspaper, Wall, Low-Texture and Cloth. We now describe each of textures used.

\textit{Bumpy} texture provides an uneven texture over the original newspaper scenario. These ``bumps'' are made of crumpled paper. The depth (protrusion) of the bumps are large and are about 25\% of the distance to gap from the quadrotor's initial position. This scenario mimics the uneven texture on rock walls around a cave opening, for example.

\textit{Leaves} texture mimics foliage. In this setup magnolia and .. leaves are glued onto foam-board. The two leaves used are of very different sizes and shapes. The leaves texture are also uneven with depth variation as large as 10\% of the distance to the quadrotor's initial position. We use both sides of the leaves. The front-side of the leaves are of a glossy texture with very high reflectance while the back-side are of matte texture with very low reflectance.  Also, the leaves look similar to each other. This texture provides similar repeating patterns and large changes in reflectance.

\textit{Door} texture is a wall with a door. Both these are white with very low texture.

\textit{Newspaper} texture is the setup similar to the one used in the main paper. Newspaper is glued onto foam-board. This presents an artificial high-texture scenario.

\textit{Wall} texture is foam-core with a small amount of logos. We consider this as a medium-texture scenario.

\textit{Low-Texture} is white foam-core with a few scribbles near the window which are required for tracking. This is artificially crafted to mimic a minimal-textured plain wall with windows.

\textit{Cloth} texture is created by the usage of wrinkled bed sheets. This scenario mimics hanging curtains, hanging paintings and hanging flags.    

A combination of the aforementioned textures creates scenarios which test the bias of the TS$^2$P algorithm.
In all the above cases, $\prescript{0}{}{Z_{\mathcal{F}}} \sim 2.8$m and $\prescript{0}{}{Z_{\mathcal{B}}} \sim 5.6$m and $N=3$ frames are used for stacking/averaging in all the cases.

\begin{figure*}[t!]
    \centering
    \includegraphics[width=\textwidth]{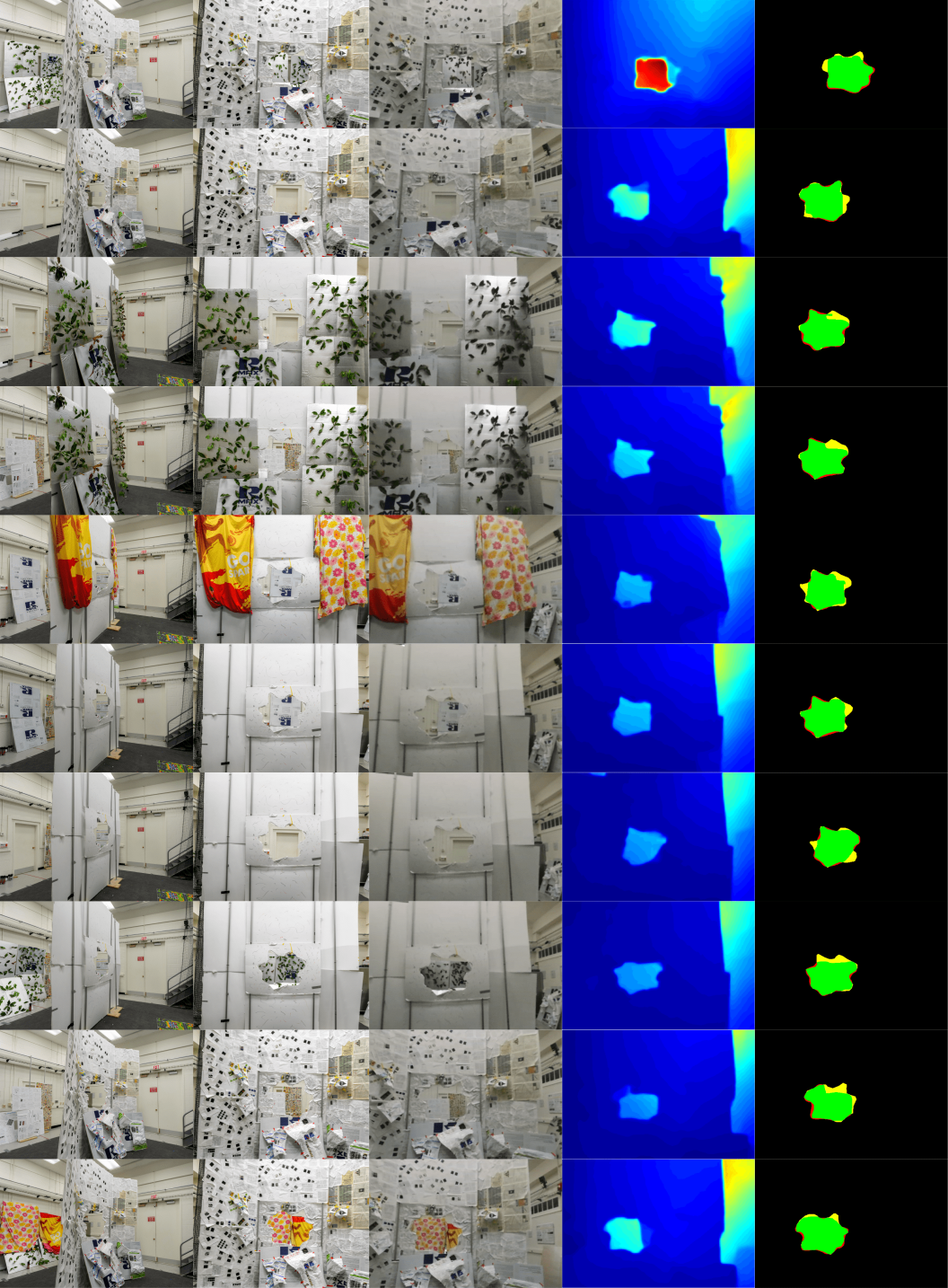}
    \caption{Left to right columnwise: Side view of the setup, Front view of the setup, sample image frame used, $\Xi$ output, Detection output - Yellow: Ground Truth, Green: Correctly detected region, Red: Incorrectly detected region. Rowwise: Cases in the order in Table \ref{tab:CasesTable}. Best viewed in color.}
    \label{fig:ExtraCases}
\end{figure*}

Our approach works in most of the scenarios as presented in Fig.~\ref{fig:ExtraCases} since it uses deep learning based optical flow to estimate the position of the gap in the image plane. Even in the low-textured scenarios, the window detection output $\mathcal{O}$ still has at least 75\% overlap with the ground truth as mentioned in our paper. TS$^2$P works even with no textures on one of the foreground or background. Though tracking the $\mathcal{F}$ and $\mathcal{B}$ without any textures is not possible.

\begin{table}[t!]
\centering
\caption{Comparison of our approach with different setups}
\resizebox{0.75\columnwidth}{!}{
\label{tab:CasesTable}
\begin{tabular}{lll}
\toprule
Scenario & Foreground & Background \\
 \hline
 1 & Bumpy & Leaves \\
 2 & Bumpy & Door  \\
 3 & Bumpy & Newspaper \\
 4 & Bumpy & Cloth \\
 5 & Leaves & Wall \\
 6 & Leaves & Newspaper \\
 7 & Cloth & Wall \\
 8 & Low-Texture & Leaves \\
 9 & Low-Texture & Leaves \\
 10 & Low-Texture & Leaves \\
 \bottomrule
\end{tabular}}
\end{table}

\end{document}